%% file: main.tex
\pdfoutput=1
\documentclass[11pt]{article}

\usepackage{EACL2023}

\usepackage{times}
\usepackage{latexsym}
\usepackage{graphicx}
\usepackage{algpseudocode}
\usepackage[T1]{fontenc}
\usepackage[utf8]{inputenc}

\usepackage{microtype}
\usepackage{inconsolata}

\usepackage{multirow}
\usepackage{amssymb}
\usepackage{amsmath}  % improve math presentation
\usepackage[linesnumbered,ruled,vlined]{algorithm2e}

\title{Analyzing Text Representations under Tight Annotation Budgets:\\ Measuring Structural Alignment}

\author{
    César González-Gutiérrez \and
    {\bf Audi Primadhanty} \and
    {\bf Francesco Cazzaro} \and
    {\bf Ariadna Quattoni} \\
    Universitat Politècnica de Catalunya, Barcelona, Spain \\
    \texttt{\{cesar.gonzalez.gutierrez, audi.primadhanty, francesco.cazzaro\}@upc.edu}
}

\begin{document}

\maketitle

\begin{abstract}
Annotating large collections of textual data can be time consuming and expensive. That is why the ability to train models with limited annotation budgets is of great importance. In this context, it has been shown that under tight annotation budgets the choice of data representation is key. The goal of this paper is to better understand why this is so. With this goal in mind, we propose a metric that measures the extent to which a given representation is \emph{structurally aligned} with a task.  We conduct experiments on several text classification datasets testing a variety of models and representations. Using our proposed metric we show that an efficient representation for a task (i.e. one that enables learning from few samples) is a representation that induces a good alignment between latent input structure and class structure.
\end{abstract}

\begin{figure*}[t]
    \centering
    \includegraphics[width=\linewidth]{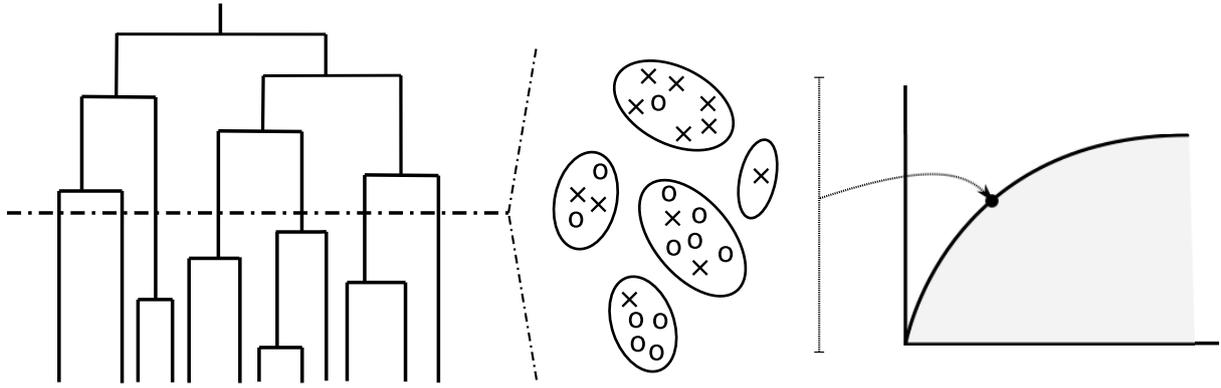}
    \caption{SAM general schema. A dendogram is first constructed from a hierarchical clustering of the representation. As we traverse the tree vertically, for each level we have a clustering. Branches correspond with clusters, merging from isolated samples (bottom) to a single group (top). For each clustering, we measure the alignment against the label classes. These scores plot a characteristic curve sweeping an area under it, which constitutes our final measure.}
    \label{fig:concept}
\end{figure*}

\section{Introduction}

With the emergence of deep learning models, the latter years have witnessed significant progress on supervised learning of textual classifiers. The caveat being that most of these methods require large amounts of training data. Annotating large collections of textual data can be time consuming and expensive. Because of this, whenever a new NLP application needs to be developed, data annotation becomes a bottleneck and the ability to train models with limited annotation budgets is of great importance. 

In this context, it has been shown that when annotated data is scarce the choice of data representation is crucial. More specifically, previous work showed that representations based on pre-trained contextual word-embeddings significantly outperform classical sparse bag-of-words representations for textual classification with small annotation budgets. This prior-work used linear-SVM models for the experimental comparisons. Our experiments further complement these conclusions by showing that the superiority of pre-trained contextual word-embeddings is true for both simple linear classifiers as well as for more complex models.  

The goal of this paper is to better understand why the choice of representation is crucial when annotated training data is scarce. Clearly, a few samples in a high dimensional space will provide a very sparse coverage of the input domain. However, if the representation space is properly aligned with the class structure even a small sample can be representative. To illustrate this idea imagine a classification problem involving two classes. Suppose that we perform a clustering on a given representation space that results on a few pure clusters (i.e. clusters such that all samples belong to the same class). Then any training set that `hits' all the clusters can be representative. Notice that there is a trade-off between the number of clusters and their purity, a well \emph{aligned} representation is one so that we can obtain a clustering with a small number of highly pure clusters. Based on this insight we propose a metric that measures the extent to which a given representation is \emph{structurally aligned} with a task. 

We conduct experiments on several text classification datasets comparing different representations. Our results show that there is a clear correlation between the structural alignment induced by a representation and performance with few training samples. Providing an answer to the main question addressed in this work: An efficient representation for a task (i.e. one that enables learning from a few samples) is a representation that induces a good alignment between latent input structure and class structure.

In summary, the main contributions of this paper are:
\begin{itemize}
    \item We show that using pre-trained word embeddings significantly improves performance under low annotation budgets, for both simple and complex models.
    \item We propose a metric to measure the extent to which a representation space is aligned with a given class structure.
    \item We conduct experiments on several textual classification datasets and show that the most efficient representations are those with an input latent structure that is well aligned to the class structure. 
\end{itemize}

The paper is organized as follows: Section \ref{related_work} discusses related work, Section \ref{representation_is_key}
presents the preliminary experiments showing that representation choice is key for good performance, Section \ref{sam} presents our proposed metric to measure representation quality, Section \ref{experiments} presents our main experimental results over four text classification datasets, finally Section \ref{conclusion} concludes the paper and discusses future work.

\section{Related Work}
\label{related_work}

The importance of representation choice has lately received a significant amount of attention from the active learning (AL) community \cite{schroder_survey_2020, zhang_active_2017}. Most of the research in AL attempts to quantify what representation is best when training the initial model for active learning, which is usually referred as the cold start problem \cite{lu_investigating_2019}. The importance of word embeddings has been also studied in the context of highly imbalanced data scenarios \cite{sahan_active_2021, naseem_comparative_2021,hashimoto_topic_2016, kholghi_benefits_2016}.

The focus of most of the research by the AL community regarding textual representations is in quantifying \emph{which} representations enable higher performance for a given task. In contrast, the focus of our paper is to understand \emph{why} a given representation performs better in a given task, with special attention to low annotation budget scenarios.

Since the objective of our contribution is to study some properties of different textual representations, this work is also related to recent work on evaluating the general capabilities of word embeddings. On this line of research, many studies are interested in testing the behaviour of such models using probing tasks that signal different linguistic capabilities \cite{conneau-etal-2018-cram, conneau-kiela-2018-senteval, marvin-linzen-2018-targeted, tenney_what_2019, miaschi_contextual_2020}. Others have targeted the capacity of word embeddings to transfer linguistic content \cite{ravishankar-etal-2019-probing, conneau-etal-2020-emerging}.

Aside from using probing tasks we now look at approaches that analyze the properties of representations directly, without an intermediate probing task. A correlation method called Singular Vector Canonical Correlation Analysis \cite{saphra-lopez-2019-understanding} has been used to compare representations during consecutive pre-training stages. Analysing the geometric properties of contextual embeddings is also an active line of work \cite{reif_visualizing_2019, ethayarajh-2019-contextual, hewitt-manning-2019-structural}.

The main differences between these works and ours is that previous work has focused on analysing geometric properties of the representations independently of a task while our focus is on studying the relationship between a representation and the labels of a downstream target task.

A recent contribution whose objective is more closely related to ours is \cite{yauney-mimno-2021-comparing}. In this work the authors present a method to measure the alignment between documents (in a given representation space) and labels for a downstream classification task based on a data complexity measure developed in the learning-theory community. The main idea is to exploit a dual representation of the documents (i.e. each document is represented by its similarity to other documents) and test for label linear separability in this space. 

There are three main differences between their work and ours: 1) Our approach does not test for linear separability (in a dual space). Instead we measure the alignment of the latent structure in the given representation space with the label class structure (potentially testing for more complex decision surfaces). 2) Their empirical study focuses on binary text classification tasks with balanced label distributions while we study both balanced and highly non-balanced scenarios. 3) Our study focuses on the low annotation budget scenario (it is in this low-budget scenario that our experiments show that representation is critical independently of the classification model).

\section{Learning Under an Annotation Budget: Choice of Representation is Key}
\label{representation_is_key}

\begin{table}[t]
\begin{tabular}{lll}
 & Samples & Labels ($-$/$+$) \\
\hline
IMDB & 50K & 25K / 25K \\
WT & 224K & 202K / 21K \\
CC & 2M & 1.84M / 160K \\
s140 & 1.6M & 800K / 800K \\
\hline
\end{tabular}
\caption{Datasets statistics with number of samples and number of labels per class: negative and positive ($-$/$+$).}
\label{table:data_stats}
\end{table}

In this section we investigate the importance of pre-trained word representations when learning with a few training samples (i.e. learning under an annotation budget). To study this, we compare several models of different levels of complexity, trained with and without pre-trained word embeddings. More specifically, we consider the following models:

\begin{itemize}
    \setlength\itemsep{0em}
    \item Max-Entropy: a standard max-entropy model trained with $l2$ regularization.
    \item WFA: this is in essence equivalent to an RNN with linear activation function \cite{quattoni-carreras-2020-comparison}.
    \item BERT: a BERT-base uncased (110M parameters) model \cite{devlin-etal-2019-bert} pre-trained on BooksCorpus and Wikipedia.
\end{itemize}

Each of the models described above will be trained in two settings: 1) with a sparse bag-of-words (BOW) representation and 2) with pre-trained word embeddings (PRE-BERT). For BERT, training with BOW means training without pre-trained word embeddings.

\begin{figure}
    \centering
\includegraphics[width=0.99\linewidth]{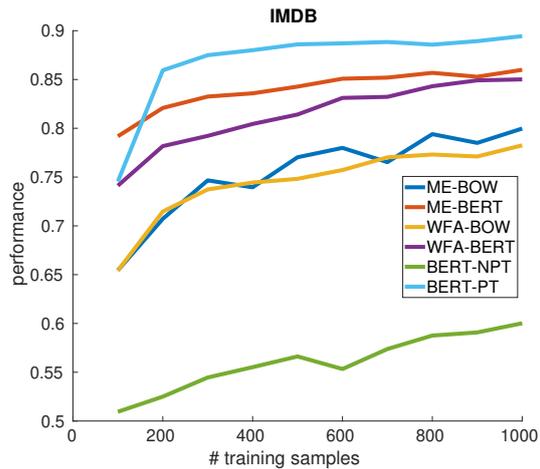}
    \caption{Performance of different models and textual representations when learning with a limited annotation budget for the IMDB dataset.}
    \label{fig:models_lc_imdb}
\end{figure}

\begin{table}
\begin{tabular}{llll}
     & & BOW & PRE-BERT \\ 
\hline 
\multirow{3}{*}{IMDB} & ME & 0.75 & 0.84 \\ 
 & WFA & 0.75 & 0.82 \\ 
 & BERT & 0.56 & {\bf 0.87} \vspace{0.5em} \\
\multirow{3}{*}{WT} & ME & 0.32 & 0.50 \\ 
 & WFA & 0.44 & 0.48 \\ 
 & BERT & 0.17 & {\bf 0.52} \vspace{0.5em} \\
\multirow{3}{*}{CC} & ME & 0.11 & {\bf 0.32} \\
 & WFA & 0.19 & 0.30 \\ 
 & BERT & 0.15 & 0.27 \vspace{0.5em} \\
\multirow{3}{*}{s140} & ME & 0.58 & {\bf 0.79} \\ 
 & WFA & 0.61 & 0.62 \\ 
 & BERT & 0.52 & {\bf 0.79} \\  
\hline 
\end{tabular}
\caption{ Model performance comparison when learning with a limited annotation budget. For each model we report the area under the learning curve when training models with 100 to 1000 training samples (ALC).}
\label{tab:models_lc}
\end{table}

For the experiments in this section and section \ref{experiments} we use four textual classification datasets with both balanced and imbalanced label distributions, covering a range of tasks and input lengths. More precisely, we run experiments on:
\begin{itemize}
    \setlength\itemsep{0em}
    \item IMDB \cite{maas-etal-2011-learning}: Movie reviews annotated with sentiment. This is a dataset with a balanced distribution of labels.
    
    \item s140 \cite{go_twitter_2009}: A collection of short messages in Twitter annotated with sentiment. This is a dataset with a balanced distribution of labels.
    
    \item WT \cite{wulczyn_ex_2017}: A collection of Wikipedia comments annotated with toxicity labels. This is a dataset with a highly unbalanced label distribution, less than \%15 of the labels correspond to toxic comments.
    
    \item CC \cite{borkan_nuanced_2019}: A collection of comments posted using Civil Comments platform annotated with respect to toxic behaviour. This is a dataset with a highly unbalanced label distribution, less than \%10 of the labels correspond to toxic behaviour.
\end{itemize}

Table \ref{table:data_stats} shows summary statistics for each dataset.

To generate a learning curve under an annotation budget, we create small training sets by selecting $N$ random samples, where $N$ ranges from $100$ to $1000$ in increments of $100$. Each model has a set of hyper-parameters to optimize, for each point $N$ in the learning curve we create an 80\%-20\% 5-fold cross validation split to find the optimal hyper-parameters. We then use these hyper-parameters to train a model using the full $N$ training samples and measure its performance on the test set.

We repeat the experiment with 5 different random seeds and report the mean results. As an evaluation metric we use: accuracy for the balanced datasets (i.e. IMDB and s140) and F1 (of the target class) for the imbalanced datasets (i.e. WT and CC). 

Figure \ref{fig:models_lc_imdb} shows the learning curve for the IMDB dataset. To summarize the learning curve results we compute a single performance number for each model, the area under the learning curve (ALC). This allows us to have a single robust metric that we can use to compare different models.

\begin{table}[t]
\begin{tabular}{llll}
& & BOW & PRE-BERT \\ 
\hline 
\multirow{3}{*}{IMDB} & ME & 0.89 & 0.89 \\ 
 & WFA & 0.86 & 0.89 \\ 
 & BERT & 0.86 & {\bf 0.93} \vspace{0.5em} \\
 
\multirow{3}{*}{WT} & ME & 0.66 & 0.61 \\ 
 & WFA & 0.66 & 0.58 \\ 
 & BERT & 0.53 & {\bf 0.68} \vspace{0.5em} \\

\multirow{3}{*}{CC} & ME & 0.60 & 0.57 \\
 & WFA & 0.56 & 0.56 \\
 & BERT & 0.20 & \textbf{0.69} \vspace{0.5em} \\ 

\multirow{3}{*}{s140} & ME & 0.81 & 0.80 \\ 
 & WFA & 0.83 & 0.82 \\ 
 & BERT & 0.77 & {\bf 0.84} \\
\hline 
\end{tabular}
\caption{ Model performance comparison with all training data.}
\label{tab:models_all_data}
\end{table}

Table \ref{tab:models_all_data} shows performance results when training with all the training data. In this setting, BERT with pre-trained word embeddings clearly outperforms all other models. However, when trained without pre-trained word embeddings, its performance is not competitive and it is outperformed by the baselines. Interestingly, BERT exhibits a large performance gain from using pre-trained word embeddings, but this is not true for the other models.

The picture is quite different when we look at low annotation budget performance (Table \ref{tab:models_lc}). In this setting, there isn't a clear winning model and both max-entropy and BERT models with pre-trained word embeddings perform similarly. But the most striking difference with respect to learning with all training data, is that with low annotation budget all models exhibit large performance gains from using pre-trained word embeddings. 

These results confirm and complement conclusions made by previous work \cite{yauney_comparing_2021} by showing that pre-trained word embeddings are useful in low annotation budget scenarios independently of the complexity of the model using them. In other words, pre-trained word-embeddings seem to be capturing some properties of the input space that can be exploited by all models. In the next section we introduce our proposed metric designed to investigate what is this main property captured by the embedding.

\section{The Structural Alignment Metric (SAM)}
\label{sam}

% self note: change SAM() for a greek letter
In this section we present the structural alignment metric (SAM) designed to measure the quality of a textual representation in the context of a given task. The idea is quite simple, in a good representation space, points that are close to each other should have higher probability of belonging to the same class. Therefore, if a representation is good, we could perform a clustering of the points and obtain clusters of \emph{high purity}, that is, clusters in which most points belong to the same class.

More precisely we assume that we are given:
\begin{itemize}
    \item A sample $S=\{(x_1,l(x_1)), \ldots, (x_n,l(x_n))\}$ of $n$ labeled points where $x \in X$ is an input point and $l(x)\in L$ is its corresponding class label. For example, for sentiment classification $x$ will be a text fragment and $l(x)$ its corresponding sentiment label, either positive or negative.
    \item A representation function $r(x): X \to \mathbb{R}^d$ mapping points in
    $X$ to some $d$-dimensional space. For example $r(x)$ can be a sparse bag-of-words representation of a text fragment.
\end{itemize}

Our goal is to compute a metric $SAM(S,r(x))$ that takes some labeled domain data and a representation function and computes a real value score.

Figure \ref{fig:concept} illustrates the steps involved in computing SAM. There are three main steps: 1) Hierarchical clustering, 2) Computing clustering partition alignments 3) Computing the aggregate metric. 

In the first step, we compute the representation of each point and build a data dendogram using hierarchical clustering. The data dendogram is built by merging clusters, progressively unfolding the latent structure of the input space. Traversing the tree, for each level we get a partition of the training points into $k$ clusters. In step 2, for each partition we measure its alignment to the class label distribution, producing an alignment curve as a function of $k$. Finally, we report the area under this curve.

\subsection{Hierarchical Clustering.}
The first step to compute SAM is to perform agglomerative clustering (implementation details are given in subsection \ref{sam_implementation}) of the points in $S$ represented using function $r(x)$. The output of this first step is a dendogram $D(S,r(x))$ (left side in figure \ref{fig:concept}) that represents a set of partitions of the points in $S$ into clusters.

More precisely, for every level of the dendogram there is an associated partition $p_k(S,r(x)):S \to \{1,\dots k\}$ that assigns each point $x$ in $S$ to one of $k$ clusters. We will regard:
$D(S,r(x))=\{p_1(S,r(x)), \ldots, p_n(S,r(x))\}$
as a set where each element is a partition of the points in $S$. To simplify notation we will write $a(p_k)$ and omit the implicit dependence on $S$ and $r(x)$.

\begin{algorithm}[ht]
\caption{Structural Alignment Metric}
\label{alg:computing_sam}
    \KwIn{\\
        $\cdot$ Data samples: \\
            $\;$ $S=\{(x_1,l(x_1)), \ldots, (x_n,l(x_n)\}$ \\
        $\cdot$ Representation: $r(x)$
    }
    \KwOut{$SAM(S,r(x))$}
    Hierarchical clustering: \newline
    We run an agglomerative clustering algorithm on the representation space defined by $r(x)$ to obtain a hierarchical dendogram: $D(S,r(x))$ of the points in $S$. \\
    Partition alignment scores: \newline
        For each partition $p_k \in D(S,r(x))$ and for each point $x \in S$ and label $l \in L$ we compute a prediction score: $z(x,l)$. Given the prediction scores and the gold labels, we compute the area under the precision recall curve (APRC) for each label, and set $a(p_k)=APRC(z,S)$ \\
    Final aggregate metric: \newline
    $SAM(S,r(x))= \int_{k_{min}}^{k_{max}} a(p_k) \,dx $ 
\end{algorithm}

\subsection{Partition Alignment Scores.}
We use $D(S,r(x))$ and the gold labels in $S$ to compute an alignment score $a(p_k)$ for each partition $p_k \in D(S,r(x))$, 

We will compute $a(p_k)$ in two steps: In the first step, we compute for every point $x \in S$ and label $l \in L$ a prediction score:
\begin{equation}
z(x,l) = \frac{\#[l,p_k(x)]}{\sum_{l\in L}\#[l,p_k(x)]}
\end{equation}
where $\#[l,p_k(x)]$ is the number of samples in cluster $p_k(x)$ with gold label $l$. Intuitively, this score assigns a label probability to a point $x$ which is proportional to the label distribution of the corresponding cluster to which $x$ belongs. 

In the second step, we use the scores $z(x,l)$ of all points and their corresponding gold labels $l(x)$ to compute the alignment score for the partition. At this point we could use several metrics to evaluate the prediction scores. 

We choose to use the area under the precision recall curve (APRC) because it has the nice property that is applicable to tasks with both balanced and unbalanced class-distributions. More specifically, for unbalanced datasets with a target positive class, we compute the APRC of the target class and for balanced datasets we compute the average of the APRC of each class (more details are given in the experimental part in Section \ref{experiments}).  

\subsection{Final Aggregate Metric: SAM}
Now that we have an alignment score for every level of the hierarchical dendogram we are ready to define our final structural alignment score (SAM). Consider the function $a(p_k)$, for the partition corresponding to the finest level of the hierarchical dendogram. The alignment score $a(p_n)$ will be $1$ because every cluster in this partition is a singleton and therefore $z(x,l)$ will be $1$ for the gold label and $0$ for all others. 

On the other extreme, for the partition corresponding to the root of the dendogram (were all points belong to a single cluster) we get that the alignment score $a(p_1)$ is the APRC corresponding to assigning to every point $x \in S $ and label $l$ the prior probability of class $l \in S$.

Consider the alignment score as a function of the size of the partition, as we increase $k$ we will get higher scores. A good representation is one that can get a high score using as few clusters as possible. Instead of choosing a fixed $k$, we believe that a more robust metric is the area under $a(p_k)$. The type of alignment that interests us is structural, and can't be expressed easily using a predefined level of granularity. 

% I COMMENTED THIS BECAUSE I DON'T UNDERSTAND IT, IT IS NOT 
% SELF EVIDENT, BETTER EXPLANATION?
%We also think this helps in alleviating the %tendency found in clustering methods of %regressing to salient features.

Now we are ready to define our final metric:
\begin{equation}
    SAM(S,r(x))= \int_{k_{min}}^{k_{max}} a(p_k) \,dx 
\end{equation}

Algorithm \ref{alg:computing_sam} summarizes the whole procedure.
%The following algorithmic box summarizes the method: 

\begin{table}[t]
\begin{tabular}{llllll}
      & IMDB & WT & CC & s140 & avg \\ 
\hline 
BOW   & 0.71 & 0.50 & 0.20 & 0.68 & 0.52 \\ 
BERTa & 0.84 & 0.67 & 0.27 & 0.75 & \bf{0.63} \\ 
BERTc & 0.80 & 0.56 & 0.22 & 0.74 & 0.58 \\
GloVe & 0.80 & 0.63 & 0.26 & 0.73 & 0.60 \\ 
FastT & 0.77 & 0.57 & 0.21 & 0.71 & 0.56  \\ 
\hline 
\end{tabular}
\caption{SAM for different representations and datasets.}
\label{tab:sam}
\end{table}

\subsection{Implementation Details.}
\label{sam_implementation}

To generate the hierarchical partitions and corresponding dendogram, we use an agglomerative clustering approach. More precisely, we use a bottom-up algorithm that starts with every sample as a singleton cluster and consecutively merges clusters according to a similarity metric and merge criterion until a single cluster is formed. We apply Ward's method \cite{ward_hierarchical_1963}, which uses  the squared Euclidean distance between samples and then minimizes the total within-cluster variance by finding consecutive pairs of clusters with the minimal increase.

The previous step produces a list of merges representing the dendogram, which we can traverse to generate a clustering partition for each value of $k$.

\begin{figure}[t]
    \centering
    \includegraphics[width=\linewidth]{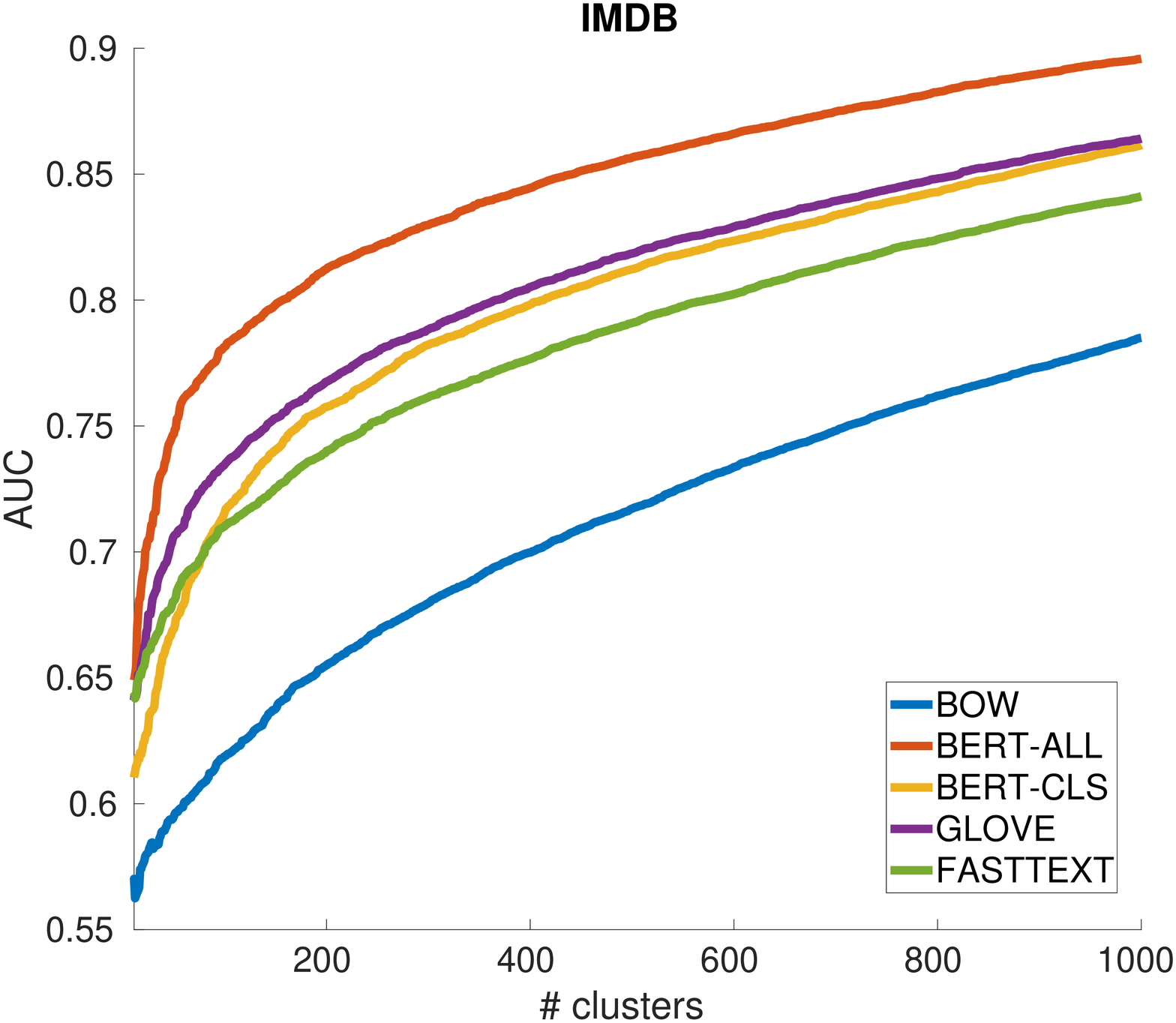}
    
    \vspace{1em}
    
    \includegraphics[width=\linewidth]{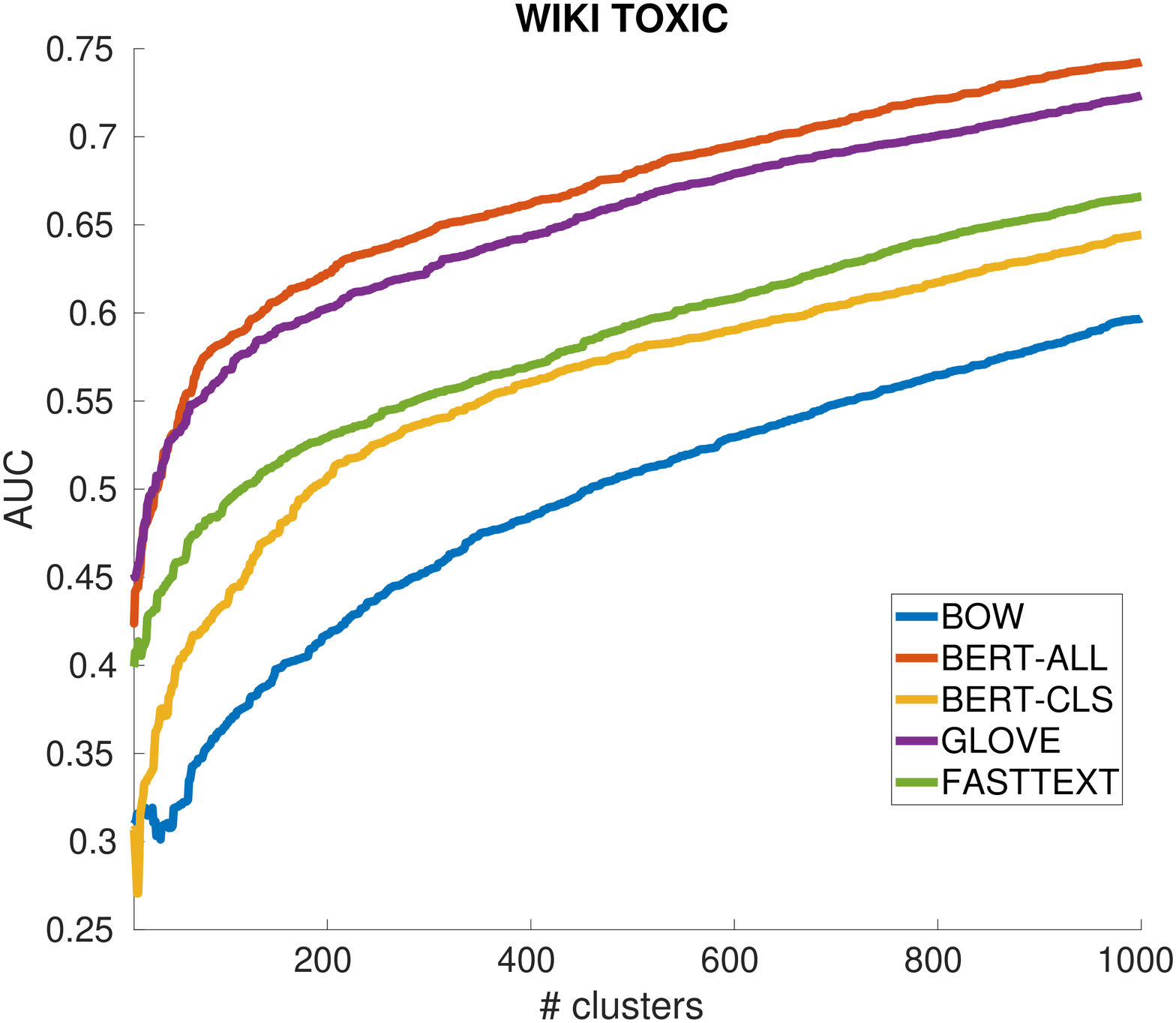}
    \caption{Alignment scores as a function of the number of clusters for IMDB and WT datasets.}
    \label{fig:sam}
\end{figure}

\section{Experimental Setup}
\label{experiments}

This section presents experiments on the four datasets described in \ref{representation_is_key}. The goal of these experiments is to analyze the correlation between the SAM of a given representation and its performance in the low annotation budget scenario. Since the focus of these experiments is on comparing representations we fix the model to be a max-entropy model. Section \ref{representation_is_key} already established that in the low annotation budget scenario max-entropy is competitive, performing similar to more complex deep-models. To generate learning curves for the different representations we follow the same experimental protocol described in \ref{representation_is_key}.

We will compare the following representations:

\begin{itemize}
    \item BOW: this is a standard sparse term frequency bag-of-words representation.
    \item \href{https://nlp.stanford.edu/projects/glove/}{GloVe}: \cite{pennington_glove_2014} word vectors pre-trained on Common Crawl (840B cased) and average pooled to get sentence vectors.
    \item FastT: \href{https://fasttext.cc/docs/en/pretrained-vectors.html}{fastText} \cite{bojanowski_enriching_2017, joulin_bag_2016} word vectors pre-trained on Wikipedia with average pooling.
    \item BERTa: \cite{devlin-etal-2019-bert} word embeddings from BERT-base uncased, 2\textsuperscript{nd} to last layer, average pooling of all tokens.
    \item BERTc: the same as above but using the \texttt{[CLS]} token alone.
\end{itemize}

\begin{table}[t]
\begin{tabular}{llllll}
& IMDB & WT & CC & s140 & avg \\ 
\hline
BOW   & 0.76  & 0.32 & 0.11 & 0.59 & 0.45 \\ 
BERTa & 0.84  & 0.50 & 0.32 & 0.79 & \bf{0.61} \\ 
BERTc & 0.80  & 0.48 & 0.23 & 0.74 & 0.56 \\
GloVe & 0.80  & 0.48 & 0.26 & 0.74 & 0.57 \\ 
FastT & 0.75  & 0.41 & 0.18 & 0.66 & 0.50 \\ 
\hline
\end{tabular}
\caption{Summary of learning curve performance (ALC) for different representations and datasets.}
\label{tab:lc}
\end{table}

We generate learning curves for each dataset and representation. Figure \ref{fig:lc} shows learning curves for IMDB and WT datasets (the other learning curves can be found in the appendix \ref{appendix:lc_sam}). We observe that BERT-a representation is consistently the best representation followed by BERT-c and GloVe performing similarly. Representations based on word-embeddings are better than the sparse baseline representation for both datasets, with the exception of FastT that doesn't exhibit a consistent improvement over BOW.

We also computed SAM for each representation and dataset. Since this metric is a measure of the alignment between a label distribution and an input representation, there is a SAM score for each label. For the balanced datasets the final number that we report is the average score over both classes, for the imbalanced datasets we report SAM over the target class (i.e. toxicity for WT and toxic-behaviour for CC). 

Figure \ref{fig:sam} shows label alignment scores as a function of the number of clusters, SAM is computed as the area under these curves. The pre-trained word embeddings, in particular BERT, tend to achieve the best results. In the curves, they show higher values of alignment for small number of clusters. BERT-a (i.e using average pooling over all tokens) seems to be superior to BERT-c (i.e. using the \texttt{[CLS]} token).

\begin{figure}[t]
    \centering
    \includegraphics[width=\linewidth]{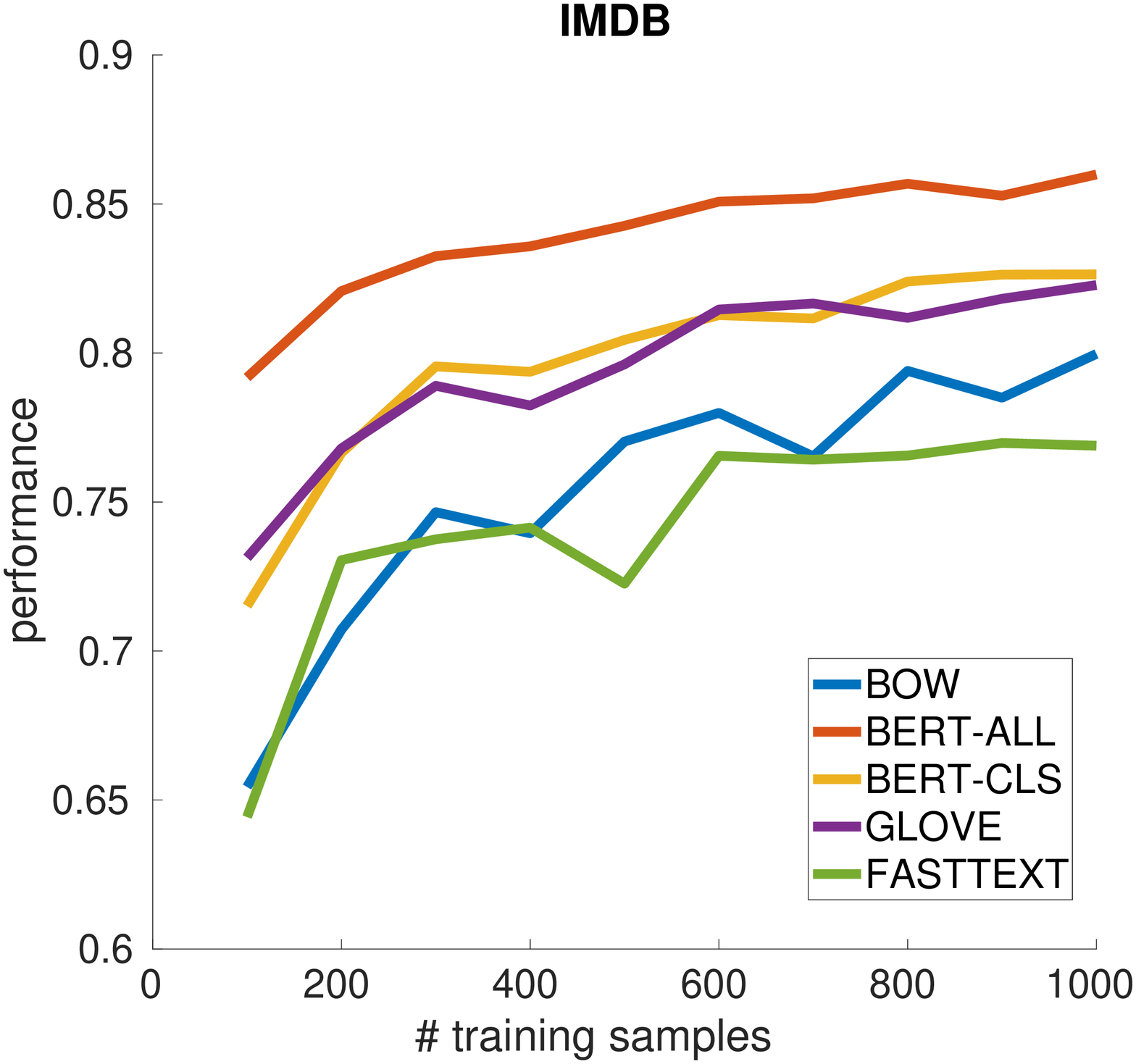}
    
    \vspace{1em}
    
    \includegraphics[width=\linewidth]{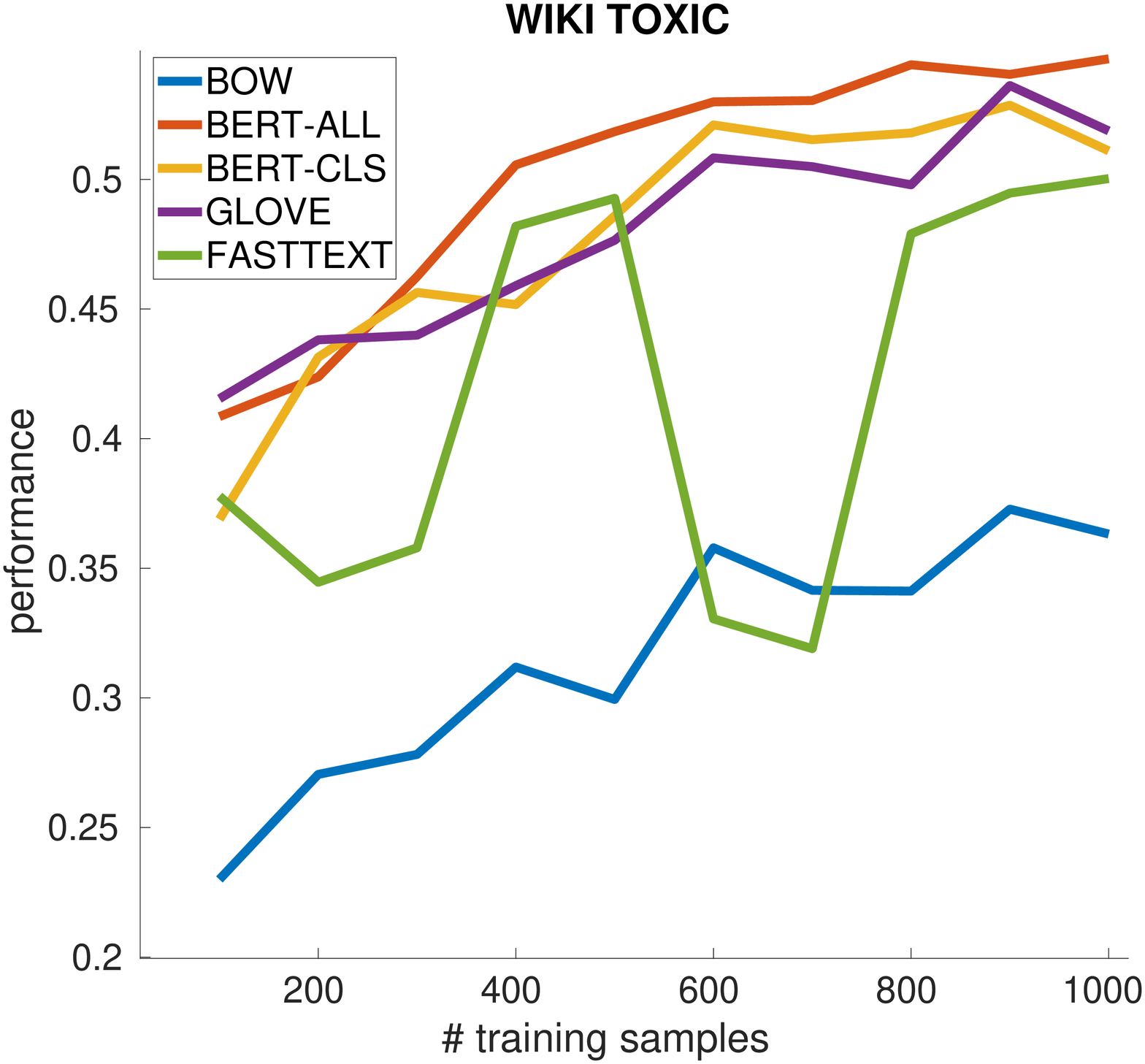}
    \caption{ Learning curve comparing different representations for IMDB and WT datasets.}
    \label{fig:lc}
\end{figure}

Tables \ref{tab:lc} and \ref{tab:sam} summarize the ALC (i.e. learning curve) and SAM results for all representations and datasets. The learning curves results show that overall BERT-a is the best representation for learning under annotation budget constraints, followed by GLoVe and BERT-c. 

All representations based on pre-trained word-embeddings outperform the baseline sparse BOW representation significantly.

When we look at the SAM scores, we see that overall it predicts accurately the relative ranking between the representations and the larger gap between the best representation (i.e. BERT-a) and the rest.

\begin{figure}[t]
    \centering
    \includegraphics[width=0.99\linewidth]{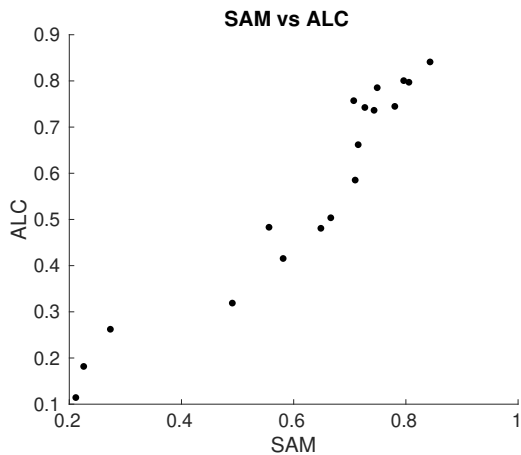}
    \caption{Correlation between SAM and Learning Curve performance.}
    \label{fig:correlations}
\end{figure}

Figure \ref{fig:correlations} shows a scatter plot of ALC as a function of SAM (each point in the plot corresponds to a dataset and representation). We observe a clear positive correlation between the two metrics.

\section{Conclusion}
\label{conclusion}

In this paper we studied the performance of different textual representations under low annotation budgets. We have shown that when labeled data is scarce, the choice of representations is key and that the use of pre-trained word embeddings improves performance significantly for all models, independently of their complexity. 

This lead us to ask the question: What is the underlying property of a good representation that is leveraged by all models in low annotation budget settings? We hypothesized that good representations are those in which the structure of the inputs is well aligned with the label distribution. We proposed a metric to measure such alignment: SAM.

To test our hypothesis we conducted experiments on several textual classification datasets, covering a variety of tasks and labeled distributions (i.e. both balanced and unbalanced). We compared a range of word-embedding representations as well as a baseline sparse representation. Our results showed that SAM is a good predictor of the performance of a given representation when labeled data is scarce.

We see several ways in which this work could be extended. In this paper we have used SAM to understand the quality of a given textual representation. However, since SAM is a function of a labeling and a representation, it could also be used to measure the quality of a labeling, given a fixed representation. For example, this might be used in the context of hierarchical labeling, to measure which level of label granularity is better aligned with some input representation. 

Another further line of work could try to generalize the notion of alignment to other tasks beyond sequence classification, such as sequence tagging.

\section*{Limitations}

The methods presented in this paper are focused on the tight annotation budget scenario. Their applicability to broader annotation scenarios could be presumed but this isn't supported by empirical results.
The experimental setup is based on binary classification tasks using English datasets. This sets the ground for further work on more complex label class structures, such as sequence tagging or ontologies, in order to asses its generality.

\section*{Acknowledgements}

The authors gratefully acknowledge the computer resources at ARTEMISA, funded by the European Union ERDF and Comunitat Valenciana as well as the technical support provided by the Instituto de Física Corpuscular, IFIC (CSIC-UV).

\bibliography{anthology,custom,zotero}
\bibliographystyle{acl_natbib}

%%\clearpage
%%\newpage
\include{appendix}

\end{document}

%% file: appendix.tex
\appendix
\section{Appendix}

\subsection{Other Learning and Alignment Curves}
\label{appendix:lc_sam}

\begin{figure*}[p]
    \centering
    \includegraphics[width=0.495\linewidth]{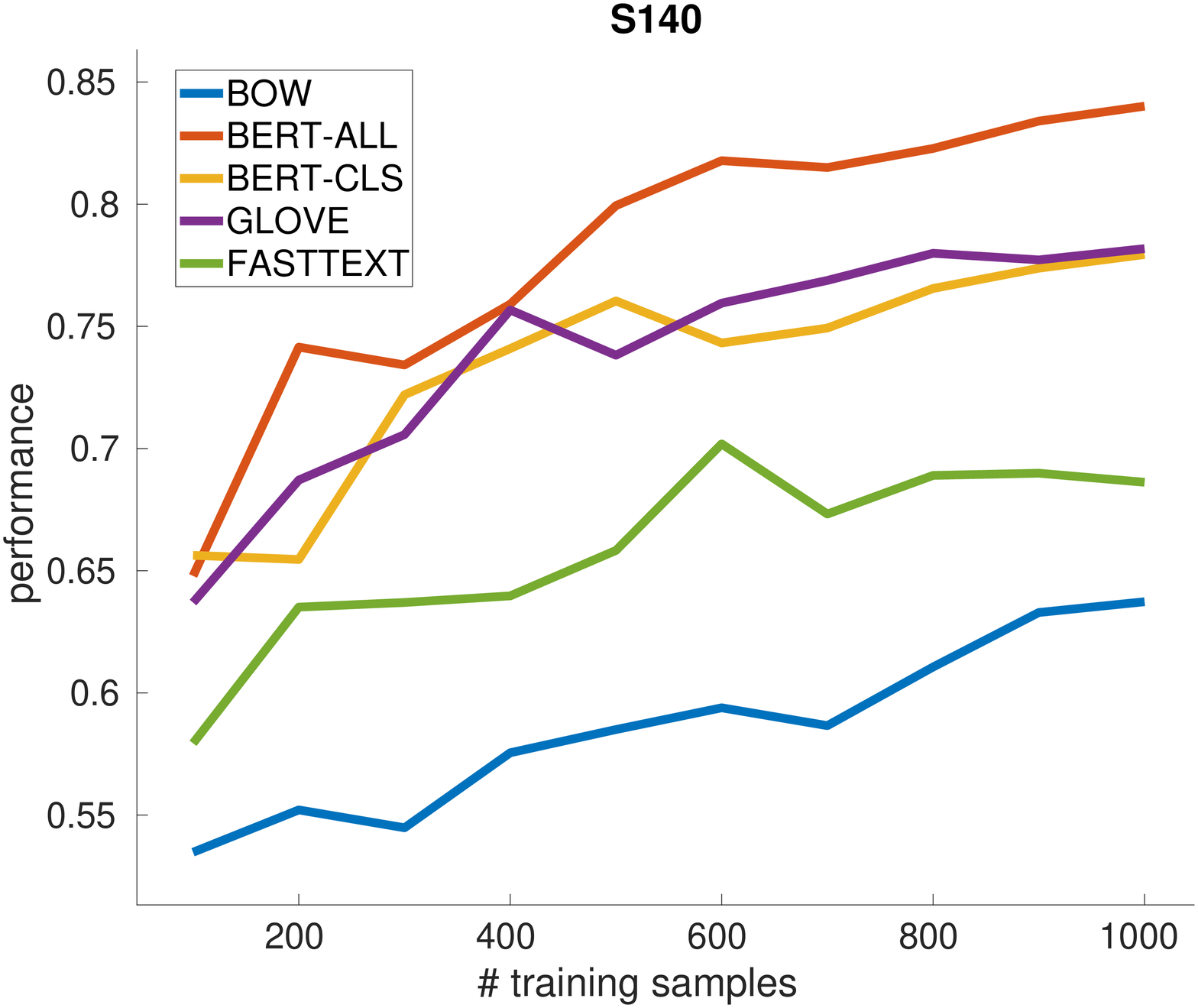}
    \includegraphics[width=0.495\linewidth]{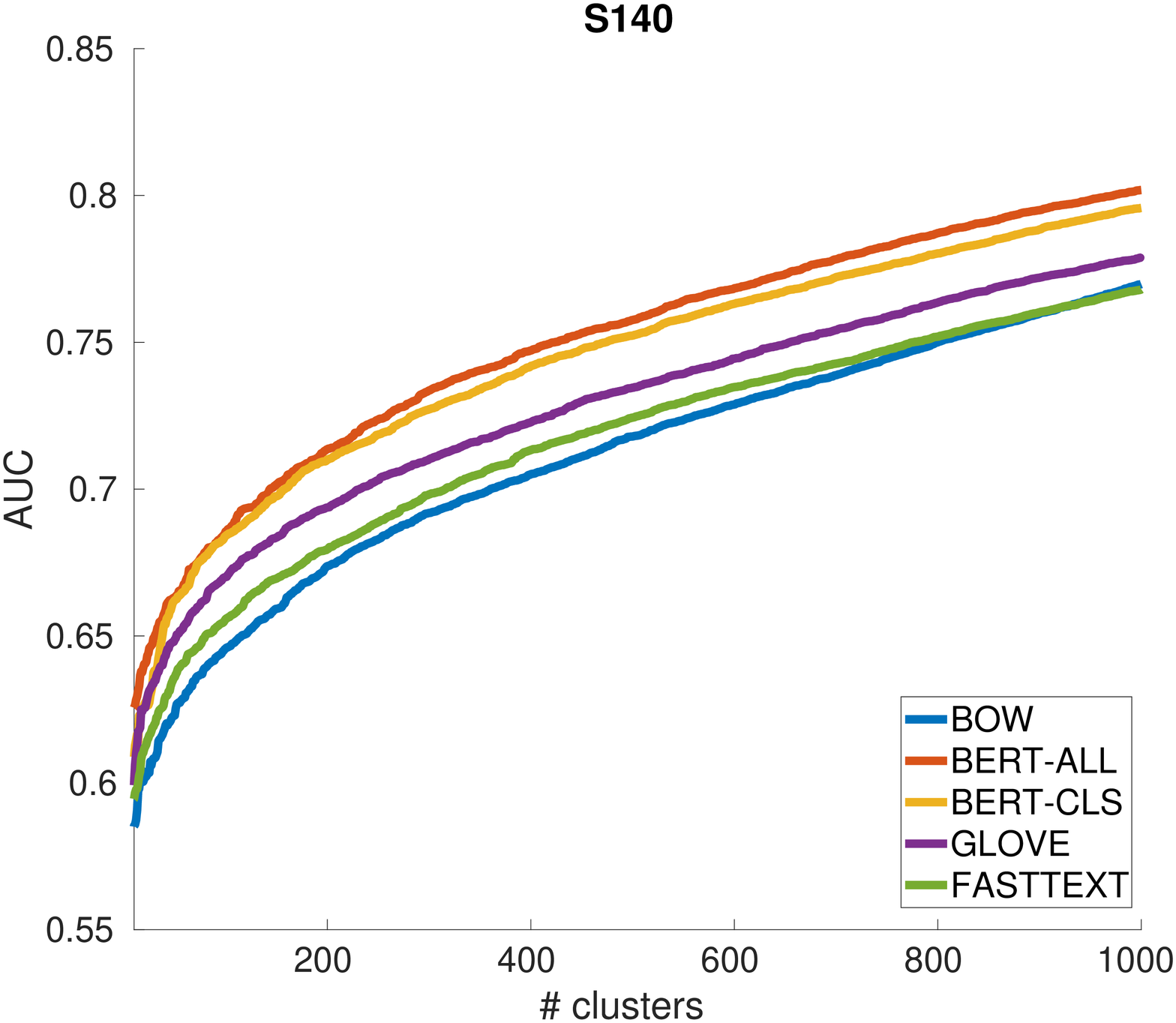}
    
    \vspace{1em}
    
    \includegraphics[width=0.495\linewidth]{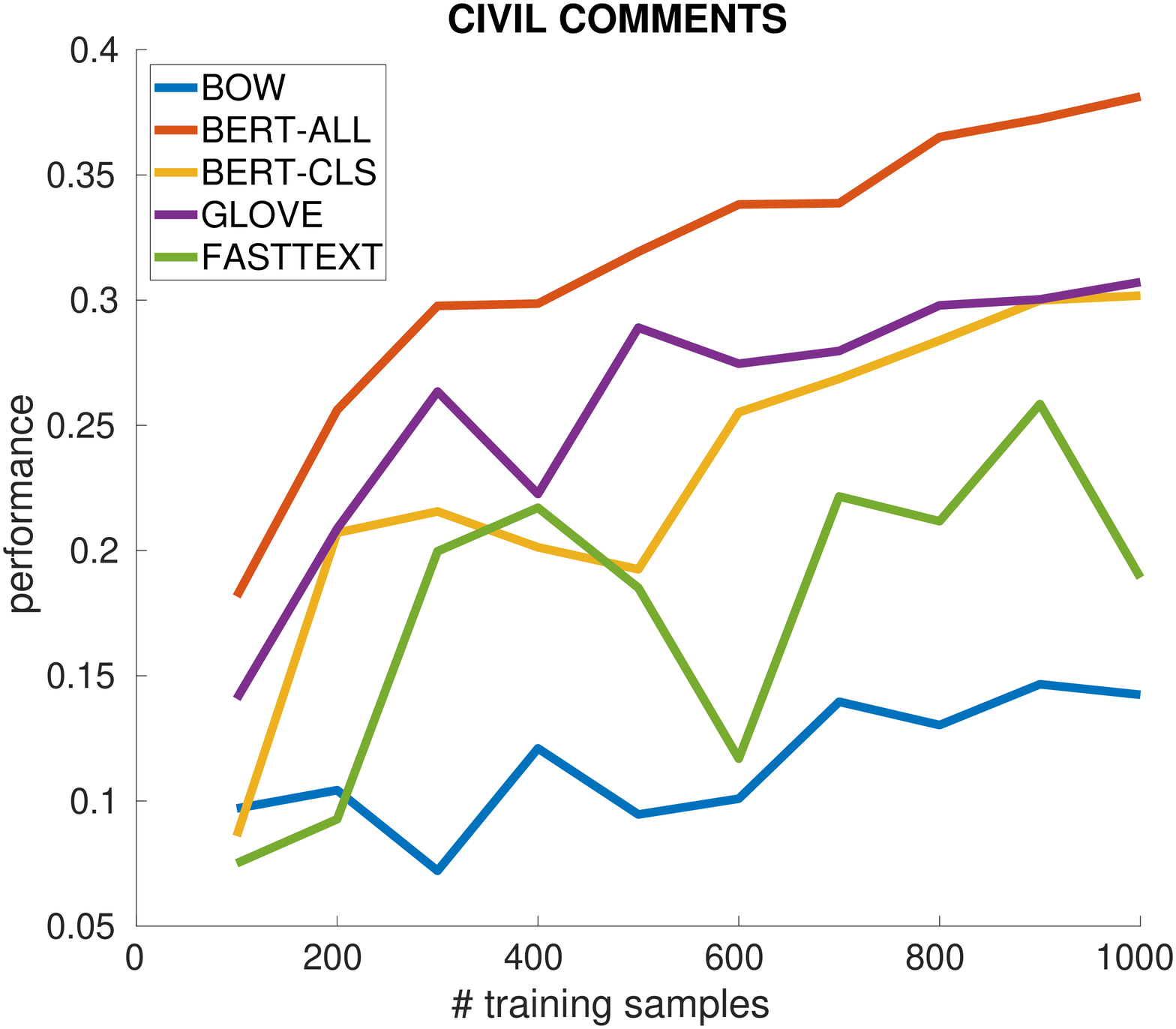}
    \includegraphics[width=0.495\linewidth]{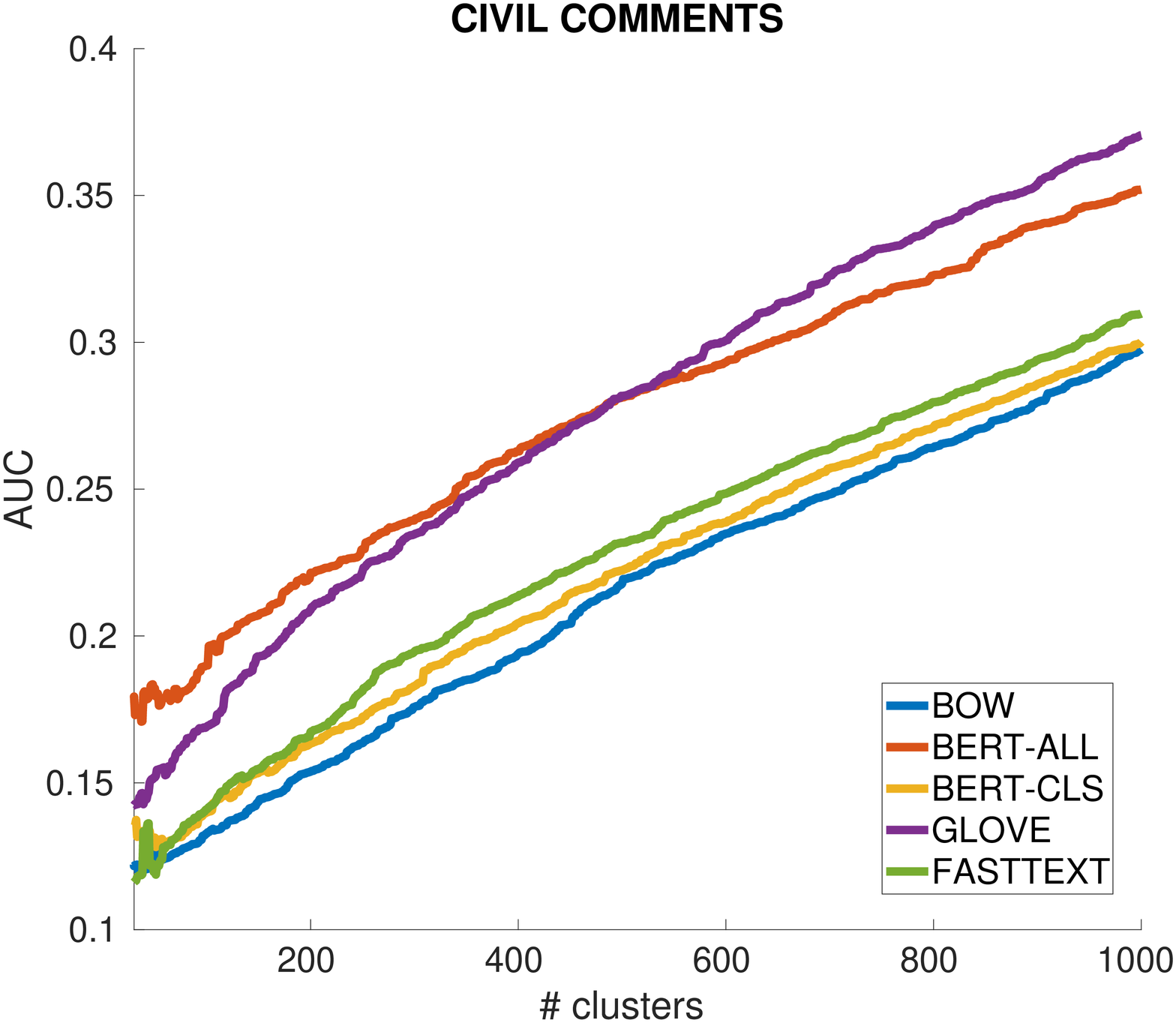}
    
    \caption{Learning curves (left) and structural alignment curves (right) for S140 and CC datasets.}
    \label{fig:lc_sam_2}
    %\vspace{7.5em}
\end{figure*}

In Figure \ref{fig:lc_sam_2} we show the learning curves and structural alignment curves for the remaining datasets introduced in Section \ref{representation_is_key}. We see a similar trend to that of the experiments presented in Section \ref{experiments}. The performance in the learning curves is paralleled in the structural alignment curves. We see a progression in performance from simpler to more sophisticated models, with all contextual representations achieving higher results with respect to the sparse bag-of-words. BERT-a performs best, with significantly better results than its relative BERT-c. The rather flat slope of Civil Comments alignment curve signals the relative difficulty of this task.

\subsection{Unsupervised Metric Curves}

For comparison with the metric developed in this work, we present similar curves and areas under the curve for an unsupervised clustering quality metric. By unsupervised we mean that only the representation space is taken into account in its computation, so the label classes aren't used at all. We have chosen to use the Davies-Bouldin index (DBI) \cite{davies_cluster_1979} to measure the quality of every clustering generated from the dendogram. Intuitively, this metric measures how compact are the clusters produced and how spread these clusters are between each other. It produces an index where lower values indicate better clusters.

\begin{figure*}[p]
    \centering
    \includegraphics[width=0.45\linewidth]{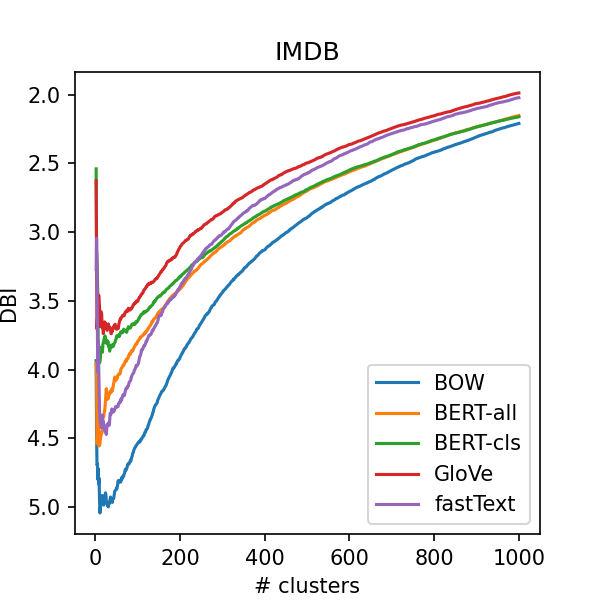}
    \includegraphics[width=0.45\linewidth]{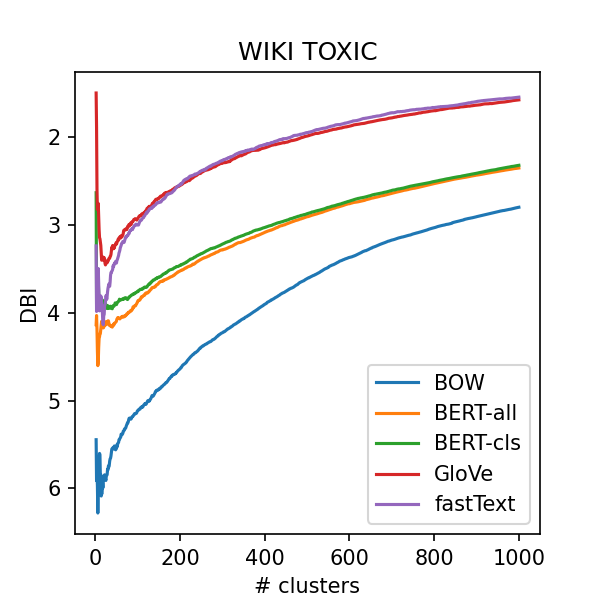}
    \includegraphics[width=0.45\linewidth]{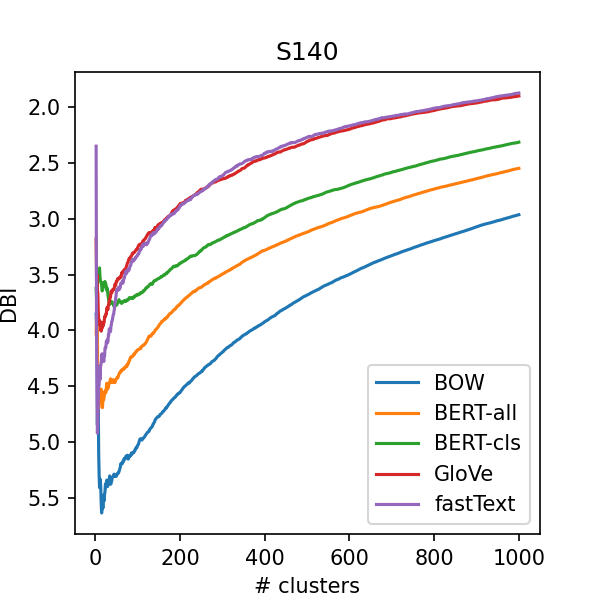}
    \includegraphics[width=0.45\linewidth]{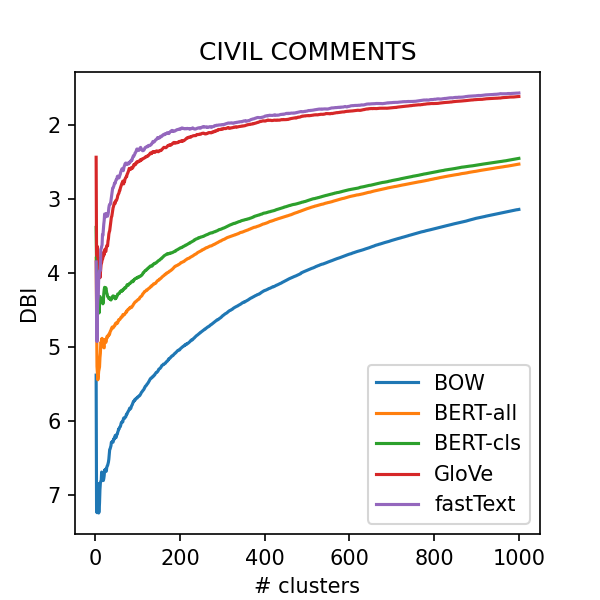}
    \caption{Davies-Bouldin index curves for different datasets. (The y-axis has been inverted for better comparison with the other curves.)}
    \label{fig:dbi}
\end{figure*}

We have computed DBI values as a function of the number of clusters to produce the curves shown in Figure \ref{fig:dbi}. As an aggregate metric, we have calculated the area under these curves. In Table \ref{tab:dbi} the reader can find a summary of the results obtained.

\begin{table}[h]
\begin{tabular}{llllll}
      & IMDB & WT & CC & s140 & avg \\ 
\hline 
BOW & 3.14 & 3.83 & 4.23 & 3.86 & 3.76 \\
BERTa & 2.87 & 3.03 & 3.31 & 3.25 & 3.11 \\
BERTc & 2.81 & 2.97 & 3.15 & 2.92 & 2.96 \\
GloVe & 2.62 & 2.12 & 2.01 & 2.47 & \textbf{2.31} \\
FastT & 2.78 & 2.13 & 1.93 & 2.47 & 2.33 \\
\hline 
\end{tabular}
\caption{Area under the Davies-Bouldin index for different representations and datasets (lower is better).}
\label{tab:dbi}
\end{table}

\begin{figure*}[p]
    \centering
    \includegraphics[width=0.45\linewidth]{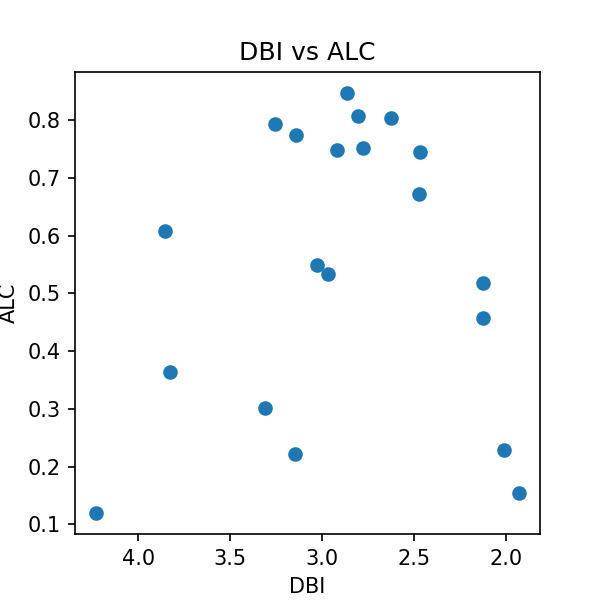}
    \caption{Correlation between DBI and ALC. Scatter plot with the area under the DBI curve against the area under the learning curve. (The x-axis has been inverted for better comparison with SAM vs ALC correlation in Figure \ref{fig:correlations}.)}
    \label{fig:dbi_vs_alc}
\end{figure*}

It is interesting to note that the best representation according to this metric, the representation that induces the best clusters, is GloVe in this case. BERT with all tokens doesn't produce particularly good clusters (according to this index), despite being the strongest representation as we showed in the experiments on learning curves in Section \ref{experiments}. To make this point clearer, we present in Figure \ref{fig:dbi_vs_alc} a scatter plot of the area under DBI with respect to the area under the learning curve for each dataset and representation. For DBI vs ALC, points are spread throughout the plot indicating low correlation. We can conclude that the quality (according to DBI) of clusters is not a good predictor of the quality of a representation in the low budget learning scenario.

\subsection{Experimental Details}

The datasets used in this paper have been extracted from \href{https://huggingface.co/datasets}{HuggingFace Datasets}. For Wikipedia toxicity detection and Civil Comments datasets, we have applied a pre-processing consisting of removing all markup code and non alpha-numeric characters except relevant punctuation.

SAM experiments have been performed over sub-samples of $10$K examples, and averaged over $5$ seeds. We run a total of $100$ experiments ($4$ datasets, $5$ representations and $5$ seeds) summing $1.2$M curve points. Each hierarchical clustering took on average $3.3$ minutes and each complete curve took around $3$ minutes to compute with $32$ CPUs and $16$GiB of RAM, whereas the DBI metric took $7.8$ hours on average with the same hardware.

BERT models were implemented using HuggingFace Transformers library \cite{wolf-etal-2020-transformers}. BERT learning curves where averaged over $5$ sub-samples and initialization seeds, and computed using $5$-fold cross validation with the number of epochs chosen over $10$. In total we performed $2400$ experiments ($4$ datasets, with and without pretraining, $5$ seeds, $10$ learning points, $5$ folds per cross validation plus $4 \times 2 \times 5 \times 10$ train and test learning points). The calculation of each learning point took about $27$ minutes on a single Nvidia V100 GPU, totalling $177$ hours of computation. BERT representations where generated with the same hardware.

In max-entropy learning curves, parameters were validated using $5$-fold cross validation and results averaged over samples drawn from $5$ seeds. This amounts to $6000$ experiments ($4$ datasets, $5$ representations, $10$ curve points, $5$ seeds and $5$ folds per validation, plus $4 \times 5 \times 10 \times 5$ train and test curve points).

%% file: main.bbl
\begin{thebibliography}{32}
\expandafter\ifx\csname natexlab\endcsname\relax\def\natexlab#1{#1}\fi

\bibitem[{Bojanowski et~al.(2017)Bojanowski, Grave, Joulin, and
  Mikolov}]{bojanowski_enriching_2017}
Piotr Bojanowski, Edouard Grave, Armand Joulin, and Tomas Mikolov. 2017.
\newblock \href {https://doi.org/10.48550/arXiv.1607.04606} {Enriching {Word}
  {Vectors} with {Subword} {Information}}.
\newblock Technical Report arXiv:1607.04606, arXiv.
\newblock ArXiv:1607.04606 [cs] type: article.

\bibitem[{Borkan et~al.(2019)Borkan, Dixon, Sorensen, Thain, and
  Vasserman}]{borkan_nuanced_2019}
Daniel Borkan, Lucas Dixon, Jeffrey Sorensen, Nithum Thain, and Lucy Vasserman.
  2019.
\newblock \href {http://arxiv.org/abs/1903.04561} {Nuanced {Metrics} for
  {Measuring} {Unintended} {Bias} with {Real} {Data} for {Text}
  {Classification}}.
\newblock \emph{arXiv:1903.04561 [cs, stat]}.
\newblock ArXiv: 1903.04561.

\bibitem[{Conneau and Kiela(2018)}]{conneau-kiela-2018-senteval}
Alexis Conneau and Douwe Kiela. 2018.
\newblock \href {https://aclanthology.org/L18-1269} {{S}ent{E}val: An
  evaluation toolkit for universal sentence representations}.
\newblock In \emph{Proceedings of the Eleventh International Conference on
  Language Resources and Evaluation ({LREC} 2018)}, Miyazaki, Japan. European
  Language Resources Association (ELRA).

\bibitem[{Conneau et~al.(2018)Conneau, Kruszewski, Lample, Barrault, and
  Baroni}]{conneau-etal-2018-cram}
Alexis Conneau, German Kruszewski, Guillaume Lample, Lo{\"\i}c Barrault, and
  Marco Baroni. 2018.
\newblock \href {https://doi.org/10.18653/v1/P18-1198} {What you can cram into
  a single {\$}{\&}!{\#}* vector: Probing sentence embeddings for linguistic
  properties}.
\newblock In \emph{Proceedings of the 56th Annual Meeting of the Association
  for Computational Linguistics (Volume 1: Long Papers)}, pages 2126--2136,
  Melbourne, Australia. Association for Computational Linguistics.

\bibitem[{Conneau et~al.(2020)Conneau, Wu, Li, Zettlemoyer, and
  Stoyanov}]{conneau-etal-2020-emerging}
Alexis Conneau, Shijie Wu, Haoran Li, Luke Zettlemoyer, and Veselin Stoyanov.
  2020.
\newblock \href {https://doi.org/10.18653/v1/2020.acl-main.536} {Emerging
  cross-lingual structure in pretrained language models}.
\newblock In \emph{Proceedings of the 58th Annual Meeting of the Association
  for Computational Linguistics}, pages 6022--6034, Online. Association for
  Computational Linguistics.

\bibitem[{Davies and Bouldin(1979)}]{davies_cluster_1979}
David Davies and Don Bouldin. 1979.
\newblock \href {https://doi.org/10.1109/TPAMI.1979.4766909} {A {Cluster}
  {Separation} {Measure}}.
\newblock \emph{IEEE Transactions on Pattern Analysis and Machine
  Intelligence}, PAMI-1:224--227.

\bibitem[{Devlin et~al.(2019)Devlin, Chang, Lee, and
  Toutanova}]{devlin-etal-2019-bert}
Jacob Devlin, Ming-Wei Chang, Kenton Lee, and Kristina Toutanova. 2019.
\newblock \href {https://doi.org/10.18653/v1/N19-1423} {{BERT}: Pre-training of
  deep bidirectional transformers for language understanding}.
\newblock In \emph{Proceedings of the 2019 Conference of the North {A}merican
  Chapter of the Association for Computational Linguistics: Human Language
  Technologies, Volume 1 (Long and Short Papers)}, pages 4171--4186,
  Minneapolis, Minnesota. Association for Computational Linguistics.

\bibitem[{Ethayarajh(2019)}]{ethayarajh-2019-contextual}
Kawin Ethayarajh. 2019.
\newblock \href {https://doi.org/10.18653/v1/D19-1006} {How contextual are
  contextualized word representations? {C}omparing the geometry of {BERT},
  {ELM}o, and {GPT}-2 embeddings}.
\newblock In \emph{Proceedings of the 2019 Conference on Empirical Methods in
  Natural Language Processing and the 9th International Joint Conference on
  Natural Language Processing (EMNLP-IJCNLP)}, pages 55--65, Hong Kong, China.
  Association for Computational Linguistics.

\bibitem[{Go et~al.(2009)Go, Bhayani, and Huang}]{go_twitter_2009}
Alec Go, Richa Bhayani, and Lei Huang. 2009.
\newblock Twitter {Sentiment} {Classiﬁcation} using {Distant} {Supervision}.
\newblock \emph{CS224N project report, Stanford}, 1(12):6.

\bibitem[{Hashimoto et~al.(2016)Hashimoto, Kontonatsios, Miwa, and
  Ananiadou}]{hashimoto_topic_2016}
Kazuma Hashimoto, Georgios Kontonatsios, Makoto Miwa, and Sophia Ananiadou.
  2016.
\newblock \href {https://doi.org/10.1016/j.jbi.2016.06.001} {Topic detection
  using paragraph vectors to support active learning in systematic reviews}.
\newblock \emph{Journal of Biomedical Informatics}, 62:59--65.

\bibitem[{Hewitt and Manning(2019)}]{hewitt-manning-2019-structural}
John Hewitt and Christopher~D. Manning. 2019.
\newblock \href {https://doi.org/10.18653/v1/N19-1419} {{A} structural probe
  for finding syntax in word representations}.
\newblock In \emph{Proceedings of the 2019 Conference of the North {A}merican
  Chapter of the Association for Computational Linguistics: Human Language
  Technologies, Volume 1 (Long and Short Papers)}, pages 4129--4138,
  Minneapolis, Minnesota. Association for Computational Linguistics.

\bibitem[{Joulin et~al.(2016)Joulin, Grave, Bojanowski, and
  Mikolov}]{joulin_bag_2016}
Armand Joulin, Edouard Grave, Piotr Bojanowski, and Tomas Mikolov. 2016.
\newblock \href {https://doi.org/10.48550/arXiv.1607.01759} {Bag of {Tricks}
  for {Efficient} {Text} {Classification}}.
\newblock Technical Report arXiv:1607.01759, arXiv.
\newblock ArXiv:1607.01759 [cs] type: article.

\bibitem[{Kholghi et~al.(2016)Kholghi, De~Vine, Sitbon, Zuccon, and
  Nguyen}]{kholghi_benefits_2016}
Mahnoosh Kholghi, Lance De~Vine, Laurianne Sitbon, Guido Zuccon, and Anthony
  Nguyen. 2016.
\newblock \href {https://aclanthology.org/U16-1003} {The {Benefits} of {Word}
  {Embeddings} {Features} for {Active} {Learning} in {Clinical} {Information}
  {Extraction}}.
\newblock In \emph{Proceedings of the {Australasian} {Language} {Technology}
  {Association} {Workshop} 2016}, pages 25--34, Melbourne, Australia.

\bibitem[{Lu et~al.(2019)Lu, Henchion, and Mac~Namee}]{lu_investigating_2019}
Jinghui Lu, Maeve Henchion, and Brian Mac~Namee. 2019.
\newblock \href {http://arxiv.org/abs/1910.03505} {Investigating the
  {Effectiveness} of {Representations} {Based} on {Word}-{Embeddings} in
  {Active} {Learning} for {Labelling} {Text} {Datasets}}.
\newblock Number: arXiv:1910.03505 arXiv:1910.03505 [cs, stat].

\bibitem[{Maas et~al.(2011)Maas, Daly, Pham, Huang, Ng, and
  Potts}]{maas-etal-2011-learning}
Andrew~L. Maas, Raymond~E. Daly, Peter~T. Pham, Dan Huang, Andrew~Y. Ng, and
  Christopher Potts. 2011.
\newblock \href {https://aclanthology.org/P11-1015} {Learning word vectors for
  sentiment analysis}.
\newblock In \emph{Proceedings of the 49th Annual Meeting of the Association
  for Computational Linguistics: Human Language Technologies}, pages 142--150,
  Portland, Oregon, USA. Association for Computational Linguistics.

\bibitem[{Marvin and Linzen(2018)}]{marvin-linzen-2018-targeted}
Rebecca Marvin and Tal Linzen. 2018.
\newblock \href {https://doi.org/10.18653/v1/D18-1151} {Targeted syntactic
  evaluation of language models}.
\newblock In \emph{Proceedings of the 2018 Conference on Empirical Methods in
  Natural Language Processing}, pages 1192--1202, Brussels, Belgium.
  Association for Computational Linguistics.

\bibitem[{Miaschi and Dell'Orletta(2020)}]{miaschi_contextual_2020}
Alessio Miaschi and Felice Dell'Orletta. 2020.
\newblock \href {https://doi.org/10.18653/v1/2020.repl4nlp-1.15} {Contextual
  and {Non}-{Contextual} {Word} {Embeddings}: an in-depth {Linguistic}
  {Investigation}}.
\newblock In \emph{Proceedings of the 5th {Workshop} on {Representation}
  {Learning} for {NLP}}, pages 110--119, Online. Association for Computational
  Linguistics.

\bibitem[{Naseem et~al.(2021)Naseem, Khushi, Khan, Shaukat, and
  Moni}]{naseem_comparative_2021}
Usman Naseem, Matloob Khushi, Shah~Khalid Khan, Kamran Shaukat, and
  Mohammad~Ali Moni. 2021.
\newblock \href {https://doi.org/10.3390/asi4010023} {A {Comparative}
  {Analysis} of {Active} {Learning} for {Biomedical} {Text} {Mining}}.
\newblock \emph{Applied System Innovation}, 4(1):23.
\newblock Number: 1 Publisher: Multidisciplinary Digital Publishing Institute.

\bibitem[{Pennington et~al.(2014)Pennington, Socher, and
  Manning}]{pennington_glove_2014}
Jeffrey Pennington, Richard Socher, and Christopher Manning. 2014.
\newblock \href {https://doi.org/10.3115/v1/D14-1162} {Glove: {Global}
  {Vectors} for {Word} {Representation}}.
\newblock In \emph{Proceedings of the 2014 {Conference} on {Empirical}
  {Methods} in {Natural} {Language} {Processing} ({EMNLP})}, pages 1532--1543,
  Doha, Qatar. Association for Computational Linguistics.

\bibitem[{Quattoni and Carreras(2020)}]{quattoni-carreras-2020-comparison}
Ariadna Quattoni and Xavier Carreras. 2020.
\newblock \href {https://doi.org/10.18653/v1/2020.sustainlp-1.21} {A comparison
  between {CNN}s and {WFA}s for sequence classification}.
\newblock In \emph{Proceedings of SustaiNLP: Workshop on Simple and Efficient
  Natural Language Processing}, pages 159--163, Online. Association for
  Computational Linguistics.

\bibitem[{Ravishankar et~al.(2019)Ravishankar, {\O}vrelid, and
  Velldal}]{ravishankar-etal-2019-probing}
Vinit Ravishankar, Lilja {\O}vrelid, and Erik Velldal. 2019.
\newblock \href {https://doi.org/10.18653/v1/W19-4318} {Probing multilingual
  sentence representations with {X}-probe}.
\newblock In \emph{Proceedings of the 4th Workshop on Representation Learning
  for NLP (RepL4NLP-2019)}, pages 156--168, Florence, Italy. Association for
  Computational Linguistics.

\bibitem[{Reif et~al.(2019)Reif, Yuan, Wattenberg, Viegas, Coenen, Pearce, and
  Kim}]{reif_visualizing_2019}
Emily Reif, Ann Yuan, Martin Wattenberg, Fernanda~B Viegas, Andy Coenen, Adam
  Pearce, and Been Kim. 2019.
\newblock \href
  {https://proceedings.neurips.cc/paper/2019/hash/159c1ffe5b61b41b3c4d8f4c2150f6c4-Abstract.html}
  {Visualizing and {Measuring} the {Geometry} of {BERT}}.
\newblock In \emph{Advances in {Neural} {Information} {Processing} {Systems}},
  volume~32. Curran Associates, Inc.

\bibitem[{Sahan et~al.(2021)Sahan, Smidl, and Marik}]{sahan_active_2021}
Marko Sahan, Vaclav Smidl, and Radek Marik. 2021.
\newblock \href {https://doi.org/10.1109/ISCSIC54682.2021.00027} {Active
  {Learning} for {Text} {Classification} and {Fake} {News} {Detection}}.
\newblock In \emph{2021 {International} {Symposium} on {Computer} {Science} and
  {Intelligent} {Controls} ({ISCSIC})}, pages 87--94. IEEE Computer Society.

\bibitem[{Saphra and Lopez(2019)}]{saphra-lopez-2019-understanding}
Naomi Saphra and Adam Lopez. 2019.
\newblock \href {https://doi.org/10.18653/v1/N19-1329} {Understanding learning
  dynamics of language models with {SVCCA}}.
\newblock In \emph{Proceedings of the 2019 Conference of the North {A}merican
  Chapter of the Association for Computational Linguistics: Human Language
  Technologies, Volume 1 (Long and Short Papers)}, pages 3257--3267,
  Minneapolis, Minnesota. Association for Computational Linguistics.

\bibitem[{Schröder and Niekler(2020)}]{schroder_survey_2020}
Christopher Schröder and Andreas Niekler. 2020.
\newblock \href {https://doi.org/10.48550/arXiv.2008.07267} {A {Survey} of
  {Active} {Learning} for {Text} {Classification} using {Deep} {Neural}
  {Networks}}.
\newblock Number: arXiv:2008.07267 arXiv:2008.07267 [cs] version: 1.

\bibitem[{Tenney et~al.(2019)Tenney, Xia, Chen, Wang, Poliak, McCoy, Kim,
  Van~Durme, Bowman, Das, and Pavlick}]{tenney_what_2019}
Ian Tenney, Patrick Xia, Berlin Chen, Alex Wang, Adam Poliak, R.~Thomas McCoy,
  Najoung Kim, Benjamin Van~Durme, Samuel~R. Bowman, Dipanjan Das, and Ellie
  Pavlick. 2019.
\newblock \href {https://doi.org/10.48550/arXiv.1905.06316} {What do you learn
  from context? {Probing} for sentence structure in contextualized word
  representations}.
\newblock Number: arXiv:1905.06316 arXiv:1905.06316 [cs].

\bibitem[{Ward(1963)}]{ward_hierarchical_1963}
Joe~H. Ward. 1963.
\newblock \href {https://doi.org/10.2307/2282967} {Hierarchical {Grouping} to
  {Optimize} an {Objective} {Function}}.
\newblock \emph{Journal of the American Statistical Association},
  58(301):236--244.
\newblock Publisher: [American Statistical Association, Taylor \& Francis,
  Ltd.].

\bibitem[{Wolf et~al.(2020)Wolf, Debut, Sanh, Chaumond, Delangue, Moi, Cistac,
  Rault, Louf, Funtowicz, Davison, Shleifer, von Platen, Ma, Jernite, Plu, Xu,
  Le~Scao, Gugger, Drame, Lhoest, and Rush}]{wolf-etal-2020-transformers}
Thomas Wolf, Lysandre Debut, Victor Sanh, Julien Chaumond, Clement Delangue,
  Anthony Moi, Pierric Cistac, Tim Rault, Remi Louf, Morgan Funtowicz, Joe
  Davison, Sam Shleifer, Patrick von Platen, Clara Ma, Yacine Jernite, Julien
  Plu, Canwen Xu, Teven Le~Scao, Sylvain Gugger, Mariama Drame, Quentin Lhoest,
  and Alexander Rush. 2020.
\newblock \href {https://doi.org/10.18653/v1/2020.emnlp-demos.6} {Transformers:
  State-of-the-art natural language processing}.
\newblock In \emph{Proceedings of the 2020 Conference on Empirical Methods in
  Natural Language Processing: System Demonstrations}, pages 38--45, Online.
  Association for Computational Linguistics.

\bibitem[{Wulczyn et~al.(2017)Wulczyn, Thain, and Dixon}]{wulczyn_ex_2017}
Ellery Wulczyn, Nithum Thain, and Lucas Dixon. 2017.
\newblock \href {https://doi.org/10.1145/3038912.3052591} {Ex {Machina}:
  {Personal} {Attacks} {Seen} at {Scale}}.
\newblock In \emph{Proceedings of the 26th {International} {Conference} on
  {World} {Wide} {Web}}, {WWW} '17, pages 1391--1399, Republic and Canton of
  Geneva, CHE. International World Wide Web Conferences Steering Committee.

\bibitem[{Yauney and Mimno(2021{\natexlab{a}})}]{yauney-mimno-2021-comparing}
Gregory Yauney and David Mimno. 2021{\natexlab{a}}.
\newblock \href {https://doi.org/10.18653/v1/2021.emnlp-main.449} {Comparing
  text representations: {A} theory-driven approach}.
\newblock In \emph{Proceedings of the 2021 Conference on Empirical Methods in
  Natural Language Processing}, pages 5527--5539, Online and Punta Cana,
  Dominican Republic. Association for Computational Linguistics.

\bibitem[{Yauney and Mimno(2021{\natexlab{b}})}]{yauney_comparing_2021}
Gregory Yauney and David Mimno. 2021{\natexlab{b}}.
\newblock \href {https://doi.org/10.18653/v1/2021.emnlp-main.449} {Comparing
  {Text} {Representations}: {A} {Theory}-{Driven} {Approach}}.
\newblock In \emph{Proceedings of the 2021 {Conference} on {Empirical}
  {Methods} in {Natural} {Language} {Processing}}, pages 5527--5539, Online and
  Punta Cana, Dominican Republic. Association for Computational Linguistics.

\bibitem[{Zhang et~al.(2017)Zhang, Lease, and Wallace}]{zhang_active_2017}
Ye~Zhang, Matthew Lease, and Byron Wallace. 2017.
\newblock \href {https://doi.org/10.1609/aaai.v31i1.10962} {Active
  {Discriminative} {Text} {Representation} {Learning}}.
\newblock \emph{Proceedings of the AAAI Conference on Artificial Intelligence},
  31(1).
\newblock Number: 1.

\end{thebibliography}
